\documentclass[conference]{IEEEtran}

\usepackage{amsmath}
\usepackage{balance}
\usepackage[noadjust]{cite}
\usepackage{gensymb}
\usepackage{threeparttable}

\ifCLASSINFOpdf
  \usepackage[pdftex]{graphicx}
\else
\fi
\usepackage{array}

\usepackage{multirow}
\usepackage{xcolor}

\begin{document}
%
\title{\huge TSV Extrusion Morphology Classification Using\\Deep Convolutional Neural Networks}



\author{\IEEEauthorblockN{Brendan Reidy\textsuperscript{1}, Golareh Jalilvand\textsuperscript{2,3}, Tengfei Jiang\textsuperscript{2,3}, Ramtin Zand\textsuperscript{1}}
\IEEEauthorblockA{\textsuperscript{1}Department of Computer Science and Engineering, University of South Carolina, Columbia, SC 29208, USA\\
\textsuperscript{2}Department of Materials Science and Engineering, University of Central Florida, Orlando, FL 32816, USA\\
\textsuperscript{3}Advanced Materials Processing and Analysis Center, University of Central Florida, Orlando, FL 32816, USA\\
e-mail: bcreidy@email.sc.edu, g\_jalilvand@knights.ucf.edu, tengfei.jiang@ucf.edu, ramtin@cse.sc.edu
}



}



%


\maketitle

\begin{abstract}
In this paper, we utilize deep convolutional neural networks (CNNs) to classify the morphology of through-silicon via (TSV) extrusion in three dimensional (3D) integrated circuits (ICs). TSV extrusion is a crucial reliability concern which can deform and crack interconnect layers in 3D ICs and cause device failures. Herein, the white light interferometry (WLI) technique is used to obtain the surface profile of the extruded TSVs. We have developed a program that uses raw data obtained from WLI to create a TSV extrusion morphology dataset, including TSV images with $54\times54$ pixels that are labeled and categorized into three morphology classes. Four CNN architectures with different network complexities are implemented and trained for TSV extrusion morphology classification application. Data augmentation and dropout approaches are utilized to realize a balance between overfitting and underfitting in the CNN models. Results obtained show that the CNN model with optimized complexity, dropout, and data augmentation can achieve a classification accuracy comparable to that of a human expert.

\end{abstract}

\begin{IEEEkeywords}
3D Integrated Circuits, Convolutional Neural Network (CNN), Data Augmentation, Deep Learning, Extrusion Morphology, Through-Silicon Via (TSV).
\end{IEEEkeywords}

%
\IEEEpeerreviewmaketitle

\section{Introduction}

\begin{figure}
\centering
\includegraphics[width=3.4in]{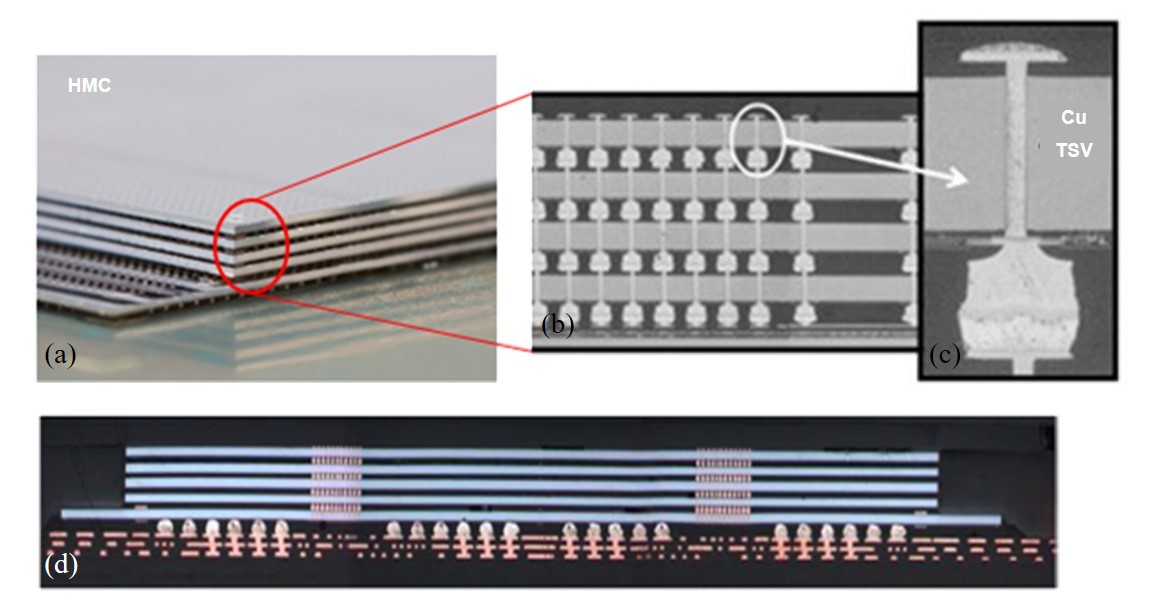}
\caption{3D integration by TSVs. (a) Scanning electron microscope (SEM) image of a hybrid memory cube (HMC) composed of a logic die at the bottom and four DRAM dies stacked on top.  (b) An SEM cross-sectional image of the HMC revealing that the stacked dynamic random-access memory (DRAM) layers are connected by vertical interconnections known as TSVs.  (c) A higher magnification image of a single TSV in a stacked die. (d) An optical microscope cross-sectional image of the entire HMC \cite{HMC}.}
\label{fig:3d integration}
\end{figure}

The semiconductor integrated circuit (IC) has profoundly impacted the electronics technology by enabling increasingly advanced products with enhanced performance and lower price \cite{mack2011fifty,havemann2001integration}. The evolution in computing and device performance driven by Moore's law has been powered by technology advancements such as the introduction of fin field-effect transistors (FinFET) and the implementation of copper (Cu)/low-k dual damascene process to continue the down-scaling of transistors. Nevertheless, basic materials and processing issues have emerged in technologies beyond 14 $nm$ node that challenge Moore's Law \cite{kuhn2009moore,hisamotofinfet,kish2002end,schulz1999end}. Thus, diligent research is being conducted on various options as potential solutions to those challenges, including three-dimensional (3D) integration. 3D integration represents a scheme in which two or more strata of dies or devices are stacked and connected through vertical interconnections, i.e. through-silicon vias (TSVs) \cite{Thomasthesis,licata1995interconnect}. In the commonly-used via-middle fabrication procedure of 3D ICs, TSVs are fabricated before the back-end-of-line (BEOL) structures through deep etching of via holes in the silicon (Si) wafer and filling of the via holes with Cu \cite{GAMBINO2015tsvfab}. An example of a TSV-based 3D IC is displayed in Figure \ref{fig:3d integration}. 

TSV-based 3D ICs offer distinct advantages including higher power efficiency, better performance, smaller form factor, and wider bandwidth, while also enabling heterogeneous integration  \cite{Knickerbocker2008,Wolf2008hetero,lau2010evolution, garrou2014handbook}. Nevertheless, during their fabrication, testing, and operation, TSVs are exposed to multiple thermal cycles, which generates considerable thermal stresses in and around the TSVs since the difference in the coefficient of thermal expansion (CTE) between Cu ($\alpha_{Cu}$=17ppm/\degree C) and Si ($\alpha_{Si}$=2.3ppm/\degree C) is very large \cite{Ryu2011,Rangan2008,slevan2009}. The thermal stress in TSVs gives rise to multiple thermo-mechanical reliability issues in the 3D ICs, including "via extrusion". Via extrusion, also known as "via protrusion", "Cu pumping" or "Cu bump", is the non-recoverable plastic deformation near the top of the via, which results in the formation of Cu bumps on the top surface of the via after thermal processing. Via extrusion is a crucial reliability concern for 3D ICs as the protrusion of Cu vias in the axial direction can deform and crack the BEOL interconnect layers to cause device failure \cite{jiang2015MRS}. Moreover, recent studies reveal that the height of via extrusion exhibits a statistical variation \cite{Jalilvand2017,jalilvand2019ectc}. The statistical nature of via extrusion is critical, as the actual reliability of a 3D IC that may contain hundreds of thousands of TSVs will be dictated by the weakest link, i.e. the 0.1\% of TSVs with the highest extrusion heights \cite{JALILVAND2019scripta}.


The variation of via extrusion can be traced to the different deformation mechanisms within the vias. Dependent upon the combined stress and microstructure states in a via, one or more mechanisms may dominate to produce extrusion of varied magnitudes \cite{jalilvand2019ectc,DeMessemaeker2014}. Statistical variation has also been observed in the morphology of via extrusion, suggesting a correlation between the extrusion mechanism and morphology \cite{jalilvand2020ectc,JALILVAND2020IEEECPMT}. Therefore, statistical study of the via extrusion morphology will provide insight on the extrusion mechanism in a large number of Cu TSVs, which is key to the development of practical solutions for the important reliability issue of via extrusion. However, studying the extrusion morphology of a large population of TSVs using conventional methods is extremely time consuming, has low throughput, and is prone to human errors. Thus, in this work, we will leverage deep learning paradigms to recognize the TSV extrusion morphology patterns and categorize them in different classes. The deep learning approaches investigated herein enable fast and accurate analyses of a large number of vias, which can reveal the statistical variation of extrusion morphology to help elucidate the extrusion mechanisms in TSV-based 3D ICs.

\section{Experimental Setup}
\label{experimental}
\begin{figure}[!b]
\centering
\includegraphics[width=0.45\textwidth]{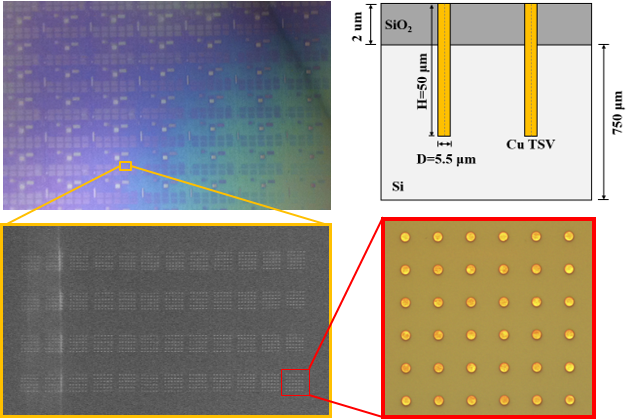}
\caption{Top and cross-sectional views of the blind via structure.}
\label{TSV schematics} 
\end{figure}

\begin{figure}
\centering
\includegraphics[width=0.45\textwidth]{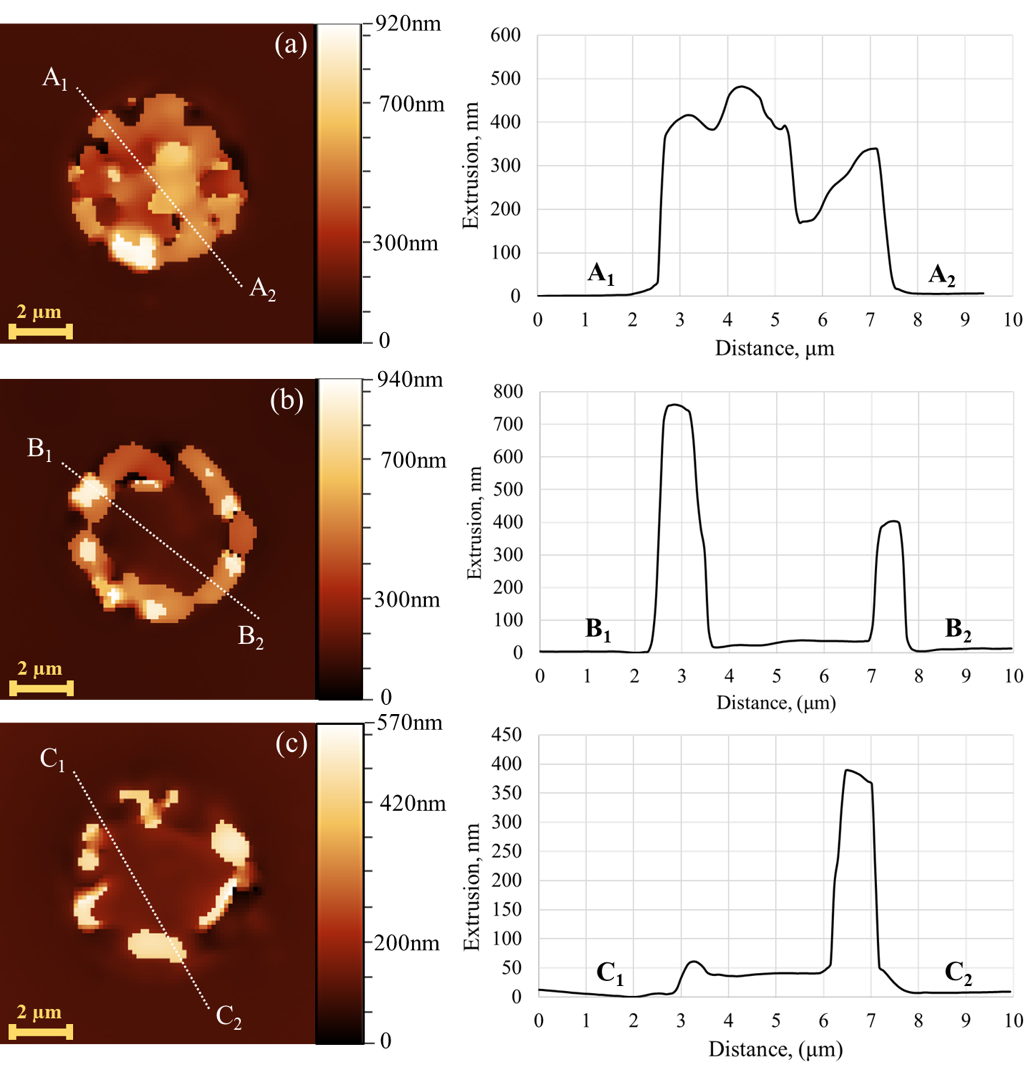}
\caption{(a) WLI surface profile and the height profile along $A_1-A_2$ in a “Granular” via. (b) WLI surface profile and the height profile along $B_1-B_2$ in an “Edge-Ring” via. (c) WLI surface profile and the height profile along $C_1-C_2$ in an “Edge-Bulge” via.}
\label{Morphologies} 
\end{figure}

The TSVs used in this study are blind vias fabricated by a standard via-middle process \cite{GAMBINO2015tsvfab} in a 750$\mu$m thick (001) Si wafer. The via dimension is 5.5$\mu$m (diameter) $\times$ 50$\mu$m (height), as illustrated in Fig. \ref{TSV schematics}. Annealing was carried out at 400\degree C for one hour in a forming gas (Ar-4\%H$_2$) atmosphere to induce extrusion. After annealing, 150$nm$ thick conformal aluminum film was deposited on the surface of the vias to enhance the surface reflectivity. Next, the surface morphology of over 2000 vias was measured by the white light interferometry (WLI) technique. The observed extrusion morphology could be categorized into three classes, as described in the following:
\begin{itemize}
    \item \textit{Granular morphology:} The extrusion occurred both in the interior and at the peripheral of the via with comparable surface undulation.
    \item \textit{Edge-Ring morphology:} The majority of extrusion occurred at the via peripheral resulting in a mostly continuous ring shape bump in the via.
    \item \textit{Edge-Bulge morphology:} The majority of extrusion occurred at the via peripheral in the form of discrete bumps.
\end{itemize}
Representative WLI images for the granular, edge-ring, and edge-bulge morphologies are shown in Fig. \ref{Morphologies}(a), (b), and (c), respectively.

\section{TSV Extrusion Morphology Dataset}

\subsection{Creating the Dataset}

\begin{figure}
\centering
\includegraphics[width=3.2in]{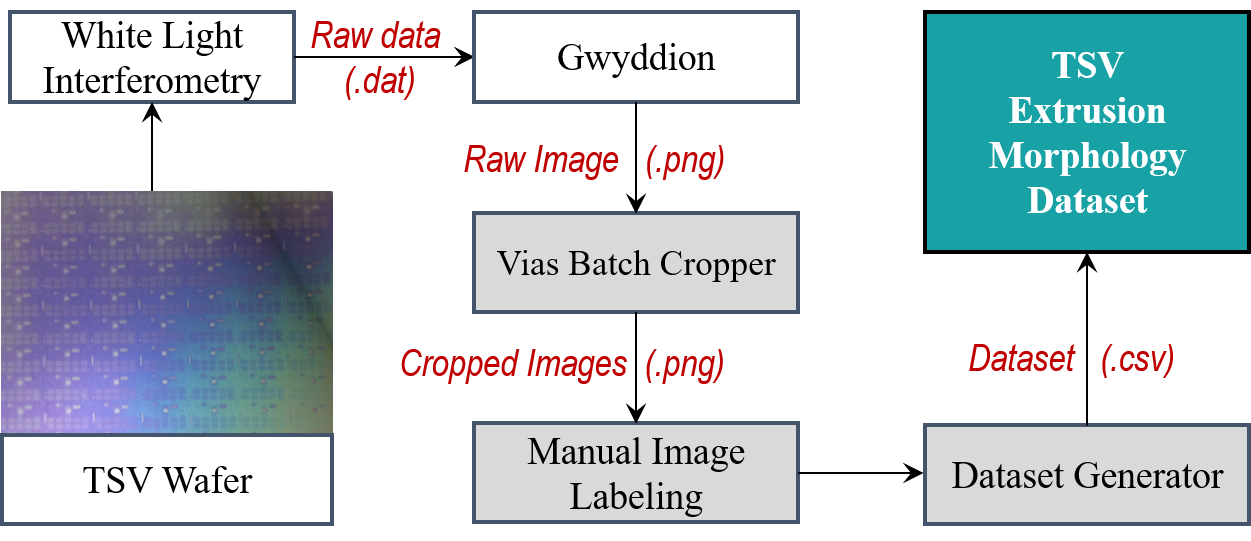}
\caption{The process of creating the TSV extrusion morphology dataset.}
\label{fig:dataset}
\end{figure}

\begin{figure}
\centering
\includegraphics[scale=1]{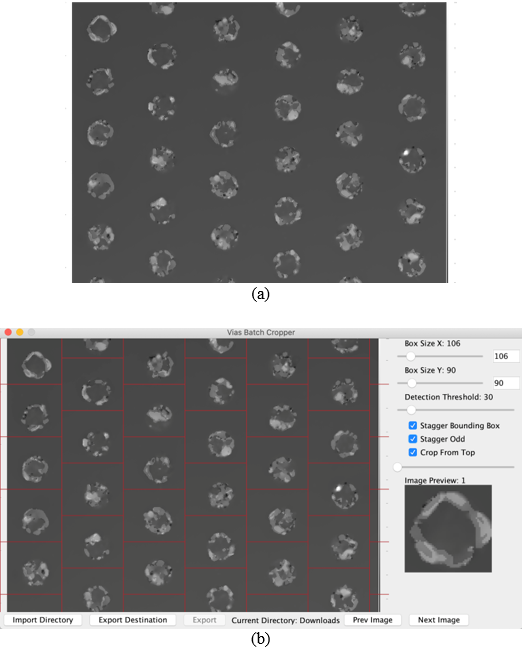}
\caption{(a) an unfiltered top-view image of TSVs obtained using WLI method and Gwyddion program. (b) a snapshot of the developed Vias Batch Cropper program used to create individual TSV images with $54\times54$ pixels from the image shown in part (a).}
\label{fig:viacropper}
\end{figure}

Figure \ref{fig:dataset} shows the process of creating the TSV extrusion morphology dataset. As described in previous section, WLI method is used to measure the surface profile of the TSVs. The raw data of the measurement results, stored in a \textit{.dat} file, is converted to unfiltered raw images (\textit{.png}) using Gwyddion \cite{Gwyddion} that is a modular program for scanning probe microscopy data visualization and analysis. An example of the raw images is shown in Fig. \ref{fig:viacropper} (a). To crop the raw images with multiple TSVs into individual images of TSVs, we developed a program called Via Batch Cropper. As shown in Fig. \ref{fig:viacropper} (b), the program uses a grid to separate TSV images, which can be fine-tuned by user to ensure each via belongs to its own cell. Since the raw images normally have a constant background color, the program can find the bounding box for each individual TSV by detecting a change in average gray-scale value over the image starting from the top left, or bottom left corner of the raw image. The program allows user to check if each image has been cropped properly using the image preview slider. If the images are not cropped properly, the user can fine-tune the cropping through the knobs provided in the program that can be seen on the right side of the Fig. \ref{fig:viacropper} (b).

Herein, we used the Via Batch Cropper program to generate over 2,000 individual images of TSVs which are fit into the center of a $54\times54$ pixel bounding box. Next, the images were sent to human experts to be annotated and manually classified into the three TSV extrusion morphology classes defined in Section \ref{experimental}. To calibrate the dataset and increase our confidence to the accuracy of labeled data, we involved multiple experts in the dataset labeling process, and repeated the process multiple times in a period of three months.

\subsection{Need for Data Augmentation}

Data augmentation plays a critical role in our work to alleviate possible over-fitting challenges which can be induced due to insufficient training data samples. As mentioned in previous sections, the TSV extrusion data acquisition process can be very time-consuming. For instance, approximately 700 person-hours were spent for data acquisition of the 2000 vias analyzed in this paper. Another challenge limiting the size of the dataset is the unbalanced distribution of the data samples across the TSV extrusion morphology classes. For example, from the 2,000 vias analyzed herein, roughly 1,200 of them belong to the granular class, while edge bulge and edge ring classes each include approximately 400 data samples. This means we could only use $\sim$400 of granular images to keep the balance between all three classes. Thus, the dataset includes a total of 1203 images with 201 and 1,002 images used for test and train phases, respectively. Authors hope that dissemination of the open-source programs developed in this work facilitates collaborations among researchers working in this field to enhance the size of the created TSV extrusion morphology dataset in the near future.

Herein, we used two geometric data augmentation techniques: \textit{rotation} and \textit{flipping}. TSV extrusion images can be rotated clockwise or counterclockwise between $1\degree$ and $359\degree$ without any changes in their labels. In addition to the rotation, we can use vertical and horizontal axis flipping techniques to further augment the size of our dataset. However, data augmentation methods such as cropping, shearing, and noise injection \cite{dataaugmentationsurvey} cannot be used to enhance the size of the TSV extrusion morphology dataset because they are not label-preserving transformations in our application. It is worth noting that the data acquisition using the WLI method is conducted in an ISO6 cleanroom and on vibration-isolating tables to avoid environment-induced noises during measurement. Thus, injecting noise to the images violates the precautions taken during the data acquisition process.


\begin{figure*}
\centering
\includegraphics[width=7in]{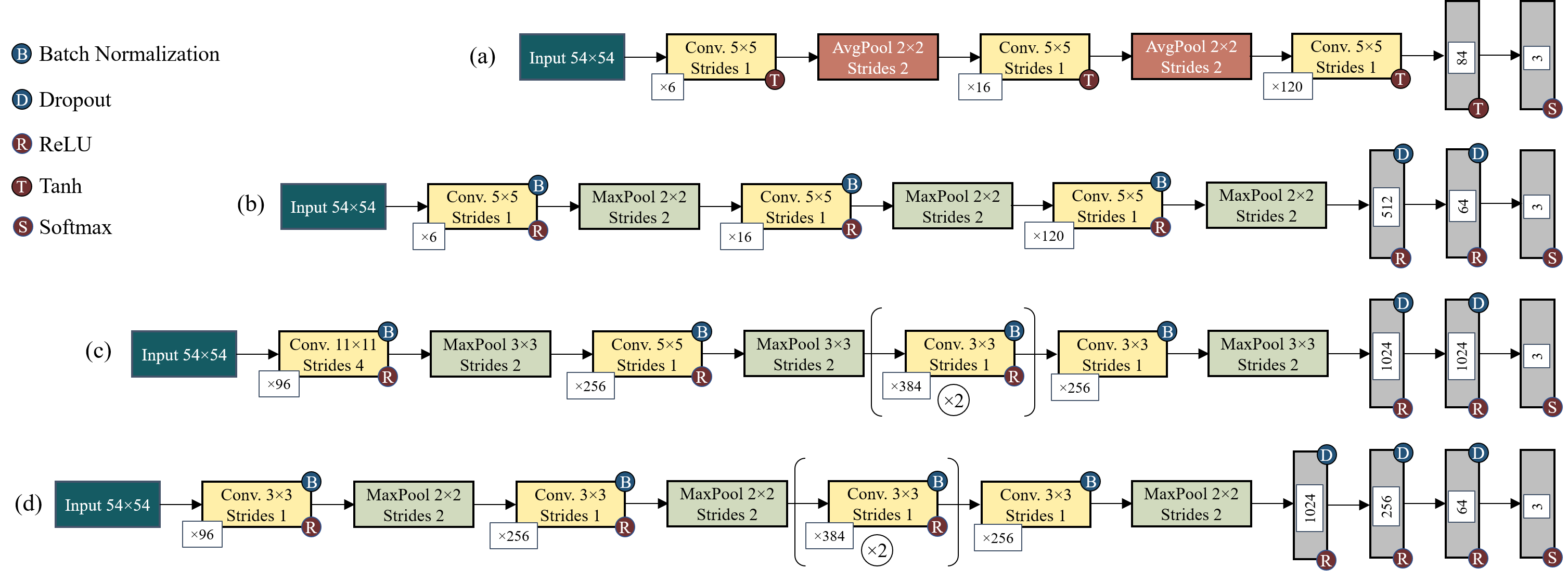}
\caption{CNNs used for TSV extrusion morphology classification: (a) LeNet-5 \cite{LeNet}, (b) AlexNet-inspired LeNet, (c) AlexNet \cite{AlexNet}, (d) VGG-inspired AlexNet.}
\label{fig:arch}
\end{figure*}

\section{CNN Architectures}
Herein, we leverage four different CNN architectures for TSV extrusion morphology classification application, as shown in Fig. \ref{fig:arch} and described below: 


\subsection{LeNet-5 \cite{LeNet}} 
The LeNet-5 architecture \cite{LeNet}, shown in Fig. \ref{fig:arch} (a), includes three convolution layers and two fully-connected layers. The first, second, and third convolution layers have 6, 16, and 120 kernels of size $5 \times 5$, respectively. The first and second convolution layers are followed by a non-overlapping average pooling layer with a filter size of $2 \times 2$ and strides of 2. The convolution layers are followed by two fully-connected (FC) layers with 84 and 3 neurons, respectively. The network uses $tanh$ activation function in all the convolution layers and the first FC layer, while the output layer uses a softmax classifier.  

\subsection{AlexNet-inspired LeNet}
Here, we propose the AlexNet-inspired LeNet architecture, shown in Fig. \ref{fig:arch} (b), which has a structure similar to LeNet-5 \cite{LeNet}, while embracing some important features from AlexNet \cite{AlexNet} CNN model. In particular, AlexNet-inspired LeNet architecture has the same number of convolution layers and kernels as LeNet-5 but all the convolution layers are followed by batch normalization and max-pooling layers instead of average pooling. Moreover, the FC layers have a $512\times64\times3$ topology with a dropout for the first and second layers. Finally, the ReLU activation function is utilized in all the convolution and fully connected layers instead of $tanh$ except for the output layer which still uses a softmax classifier.

\subsection{AlexNet \cite{AlexNet}}
The AlexNet CNN architecture \cite{AlexNet}, shown in Fig. \ref{fig:arch} (c), includes five convolution layers and three fully-connected layers. The first convolutional layer filters the $54\times54$ input TSV image with 96 kernels of size $11\times11$ with a stride of 4 pixels. The second convolutional layer includes 256 kernels of size $5 \times 5$ with a stride of 1 pixel. The first and second convolution layers are followed by batch normalization and overlapping max-pooling layers with a filter size of $3 \times 3$ and strides of 2. The third, fourth, and fifth convolution layers have 384, 384, and 256 kernels of size $3 \times 3$, respectively. The fifth convolution layer is connected to a max-pooling layer with a filter size of $3 \times 3$ and a stride of 2. The convolution layers are followed by fully connected layers with $1024\times1024\times3$ topology. Dropout is used in the first two fully-connected layers. The network uses ReLU activation function in all the convolution and fully connected layers except the output layer which uses a softmax classifier.      

\subsection{VGG-inspired AlexNet}
Here, we propose VGG-inspired AlexNet architecture which leverages some of the features used in VGG-16 CNN model \cite{VGG-16} in an AlexNet architecture, as shown in Fig. \ref{fig:arch} (d). In particular, we used very small convolution kernels of size $3 \times 3$, non-overlapping max pooling with the filter size of $2 \times 2$, and a stride of 2, and four FC layers with a $1024\times256\times64\times3$ topology. Using the above CNN model each $54\times54$ input image is converted to 256 feature maps of size $3\times3$ before being flattened and fed to the FC layers. It is worth noting that the proposed CNN architecture does not require padding in any of its layers.

\section{Results and Discussion}
Herein, we use TensorFlow \cite{tensorflow} platform and Keras library \cite{keras} to implement and optimize different CNN models for TSV extrusion morphology classification application, as described in the following:

\subsection{CNN Model Optimization to Avoid Overfitting/Underfitting}
A machine learning model suffers from the underfitting problem when it cannot capture the underlying trend of data, which is normally caused when there are not sufficient training samples available. On the other hand, overfitting deficiency occurs when a model learns the noise and random fluctuations in the training data that cannot be generalized to new data. This negatively impacts the generalization ability of the model and consequently hurts its accuracy on new data.

\begin{figure}[]
\centering
\includegraphics[width=3.4in]{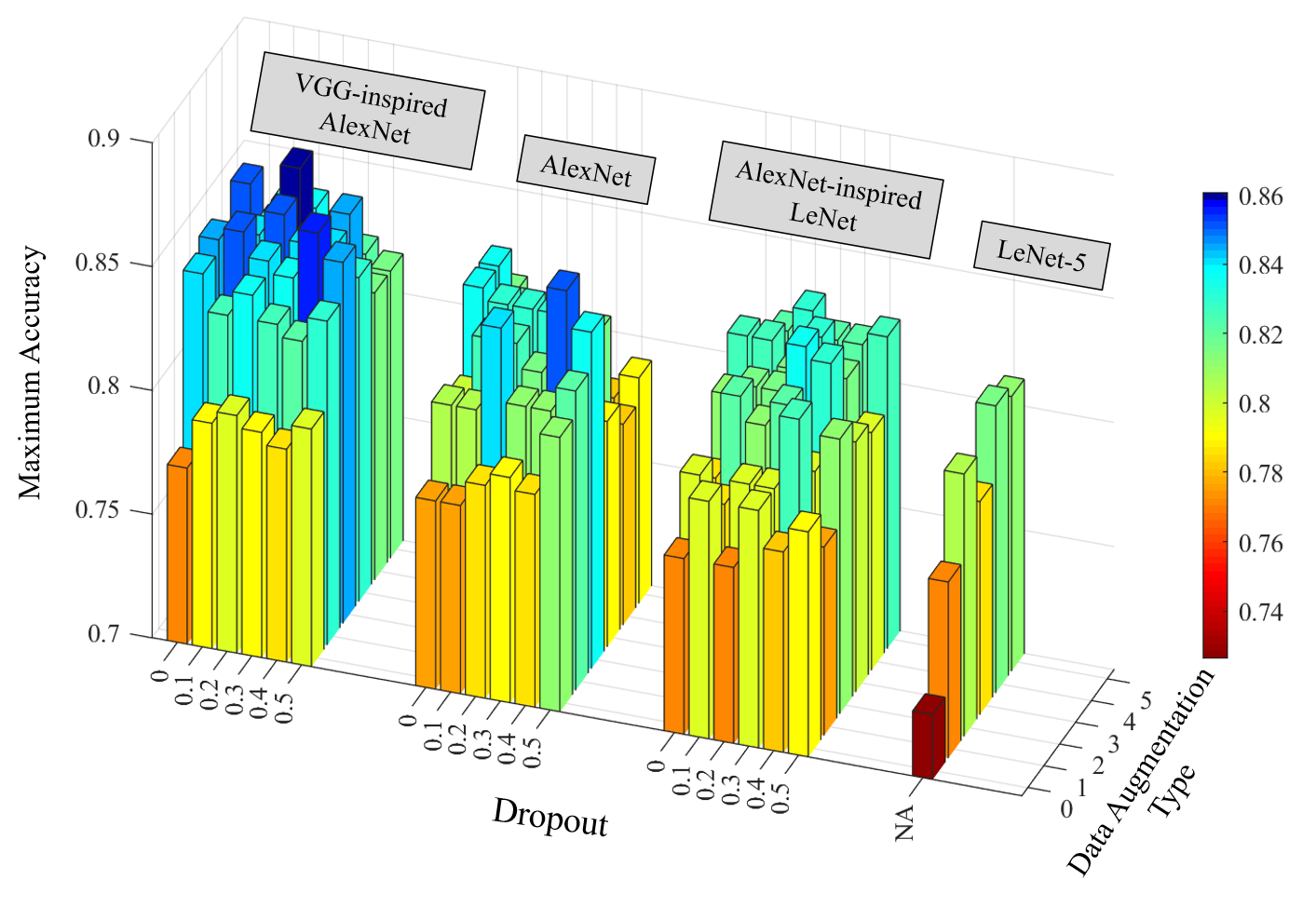}
\caption{A comparison among maximum classification accuracy values realized by four CNN models investigated herein using various data augmentation types and dropout values.}
\label{fig:maxaccu}
\end{figure}

Some of the parameters that can be optimized to achieve a balance between underfitting and overfitting are: \textit{network complexity}, \textit{data augmentation} \cite{dataaugmentationsurvey}, and \textit{dropout} \cite{dropout}. Herein, to find the optimized CNN model for our targeted application, we conducted a comprehensive experiment on the CNN models introduced in the previous section, each of which has a different network complexity. First, five different-sized augmented train datasets were developed using the rotating and flipping data augmentation methods, as listed in Table \ref{tab:dataaug}. Next, each of the four CNN models is trained for 200 epochs using various augmented datasets with dropout values ranging from 0 to 0.5. Figure \ref{fig:maxaccu} shows the maximum accuracy realized by various CNN architectures using different augmentation types and dropout values. The results obtained show that the proposed VGG-inspired AlexNet architecture can achieve the best classification accuracy of 86.1\% using Type-4 data augmentation and dropout value of 0.2. Table \ref{tab:bestmodels} lists the maximum accuracy values achieved by each CNN architecture and their corresponding hyperparameters. It is worth noting that we also performed experiments with dropout values ranging from 0.6 to 0.9, as well as larger-sized augmented datasets, however, we could not obtain accuracies higher than those achieved and shown in Figure \ref{fig:maxaccu}.

\begin{table}[]
\caption{Various data augmentation types used in this work.}
\footnotesize
\centering
\begin{tabular}{cccc}
\hline
\begin{tabular}[c]{@{}c@{}}Type\end{tabular} & \begin{tabular}[c]{@{}c@{}}Rotation Degree$\degree$\end{tabular} & \begin{tabular}[c]{@{}c@{}} Flipping\end{tabular} & \begin{tabular}[c]{@{}c@{}}Train Dataset Size\end{tabular} \\ \hline
0                                                                 & 0                                                         & NO                                                                     & 1004                                                          \\
1                                                                 & 0                                                         & YES                                                                    & 3012                                                          \\
2                                                                 & 90,180,270                                                  & NO                                                                     & 4016                                                          \\
3                                                                 & 90,180,270                                                   & YES                                                                    & 6024                                                          \\
4                                                                 & 45,90,135,180,225,270,315                                                   & NO                                                                     & 8032                                                          \\
5                                                                 & 45,90,135,180,225,270,315                                                & YES                                                                    & 10040                                                          \\ \hline
\end{tabular}
\label{tab:dataaug}
\end{table}

\begin{table}[!t]
\caption{Maximum classification accuracy achieved by studied CNN models and their corresponding hyperparameters.}
\begin{threeparttable}[b]
\begin{tabular}{lccc}
\hline
CNN Model              & \begin{tabular}[c]{@{}c@{}}Data Augmentation\\ Type\end{tabular} & Dropout & \begin{tabular}[c]{@{}c@{}}Maximum\\ Accuracy\end{tabular} \\ \hline
LeNet-5 \cite{LeNet}               & 4                                                                & NA\tnote{*}      & 81.6\%                                                     \\
AlexNet inspired LeNet & 3                                                                & 0.3     & 83.6\%                                                     \\
AlexNet \cite{AlexNet}                & 2                                                                & 0.4     & 85.1\%                                                     \\
VGG-inspired AlexNet   & 4                                                                & 0.2     & 86.1\%                                                     \\ \hline
\end{tabular}
\begin{tablenotes}
\small
\item[*] \footnotesize LeNet-5 \cite{LeNet} model does not include dropout (refer to Fig. \ref{fig:arch} (a)).
\end{tablenotes}
\end{threeparttable}
\label{tab:bestmodels}
\end{table}


\begin{table}[]
\centering
\caption{Per-class accuracy for the optimized CNN models.}
\begin{tabular}{lcccc}
\hline
\multicolumn{1}{c}{CNN Model} & \begin{tabular}[c]{@{}c@{}}Edge \\ Bulge\end{tabular} & Granular & \begin{tabular}[c]{@{}c@{}}Edge \\ Ring\end{tabular} & \multicolumn{1}{l}{\textbf{Total}} \\ \hline
LeNet-5 \cite{LeNet}                    & 80.1\%                                                & 76.1\%   & 88.1\%                                               & \textbf{81.6\%}                    \\
AlexNet inspired LeNet        & 88.1\%                                                & 77.6\%   & 85.1\%                                               & \textbf{83.6\%}                    \\
AlexNet \cite{AlexNet}                       & 76.1\%                                                & 86.6\%   & 92.5\%                                               & \textbf{85.1\%}                    \\
VGG-inspired AlexNet          & 82.1\%                                                & 88.1\%   & 88.1\%                                               & \textbf{86.1\%}                    \\ \hline
\end{tabular}
\label{tab:perclass}
\end{table}

\begin{figure}[]
\centering
\includegraphics[width=3.4in]{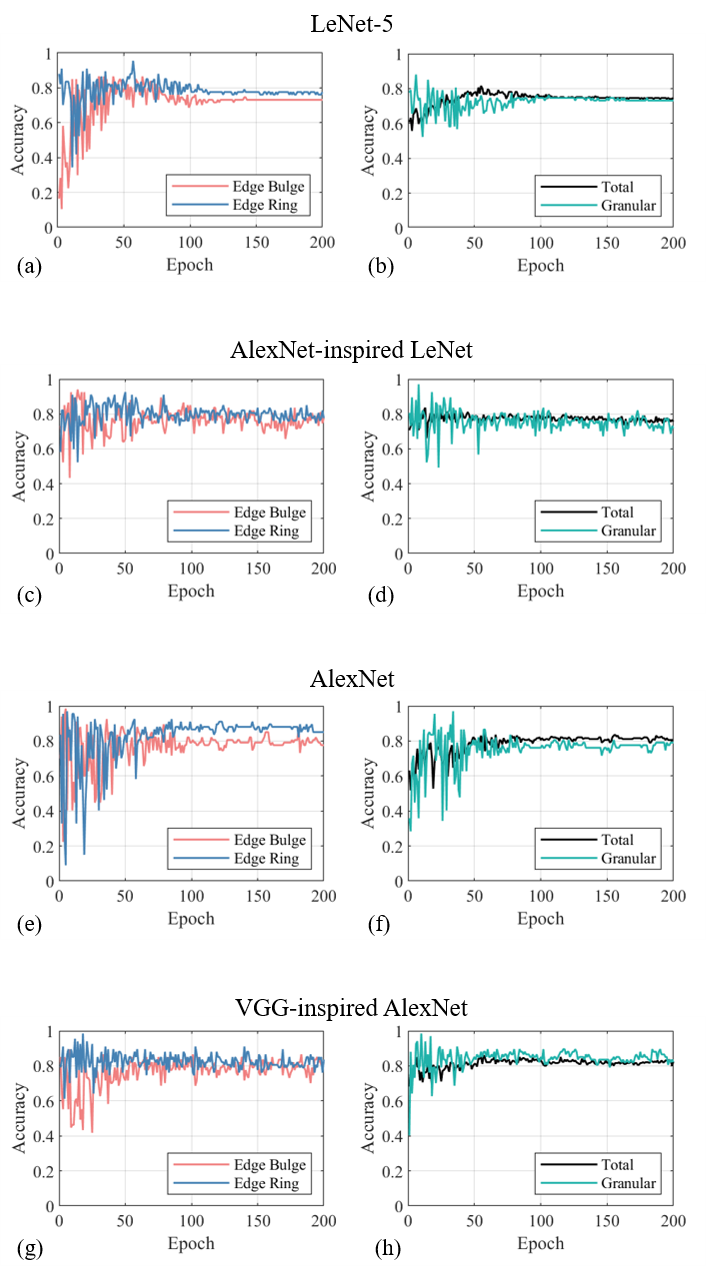}
\caption{Per-class accuracy for four optimized CNN models investigated herein. (a), (c), (e), and (g) comparison between the edge bulge and edge ring classification accuracy. (b), (d), (f), and (h) comparison between granular class and total dataset accuracy.}
\label{fig:classaccu}
\end{figure}

\subsection{Per-class Accuracy}
Here, we investigate the per-class accuracy for each of the four CNN architectures introduced in the previous section. Table \ref{tab:perclass} lists the per-class accuracy values for the optimized CNN models listed in Table \ref{tab:bestmodels}. Moreover, Fig. \ref{fig:classaccu} shows the per-class and total accuracy of the optimized CNN models in 200 training epochs. In particular, Fig. \ref{fig:classaccu} (a), (c), (e), and (g) exhibit a comparison between edge bulge and edge ring classification accuracy, while Fig. \ref{fig:classaccu} (b), (d), (f), and (h) provide a comparison between granular class accuracy and total accuracy for various CNN models studied herein. The results obtained exhibit an inverse relationship between the edge bulge and edge ring class accuracies. It means that in most of the epochs that the accuracy of edge bulge class peaks, the edge ring class's accuracy drops, and vice versa. Moreover, as shown in Fig. \ref{fig:classaccu} and listed in Table \ref{tab:perclass}, the proposed VGG-inspired AlexNet model shows the best accuracy in classifying images that belong to granular morphology, while the edge bulge and edge ring class accuracies are relatively comparable among the four studied CNN models.

\subsection{Discussion}
In addition to the current limitations in the size of the dataset, another challenge constraining the accuracy of TSV extrusion morphology classification is the data complexity. In other words, occasionally it is not possible to categorize some of the TSV samples to only one class. Here, we have conducted an experiment to better recognize this challenge. After months of dataset labeling calibration, we asked a human expert to manually-classify the same test images used for evaluating the four CNN models described in the previous sections. The results showed an 89.1\% accuracy for the expert which is comparable to the 86.1\% maximum accuracy achieved by the optimized VGG-inspired AlexNet model. This mainly occurs due to the fact that some data samples belong to two or in some cases all three morphology classes. This is also verified in laboratory experiments, based on which more than one mechanism could be involved in TSV extrusion leading to more than one extrusion morphology being observed simultaneously. Table \ref{tab:complextsvs} shows four examples of TSV extrusion images which can belong to more than one morphology classes. For instance, while sample (d), shown in the last row of the table, was originally labeled as an edge bulge morphology, edge ring and granular morphologies can also be selected as second and third morphology choices, respectively, implying that more than one mechanisms were involved in the TSV extrusion.

One possible solution to address the data complexity challenge is modifying the labeling process and assigning the controversial images to multiple classes with various degrees of confidence. However, in that case, a CNN model would require greater number of data samples and larger train dataset to learn these complex associations. Authors hope that publishing the outcome of this work and releasing the open-source programs developed herein to build the TSV extrusion morphology dataset encourage the researchers working in the field to contribute to the pool of data, which can enable several possibilities for the future work. The dataset, programs, and codes developed in this project are available for the research community in the project's GitHub repository\footnote{https://github.com/iCAS-Lab/Deep-Morphology}.

\begin{table}[]
\centering
\caption{Examples of TSV extrusion images, the morphology of which can belong to more than a single class.}
\begin{tabular}{lcl}
\hline
Sample Image & Originally-Labeled Class & \multicolumn{1}{c}{Possible Classes}                                               \\ \hline
(a) \parbox[c]{5em}{\includegraphics[width=4em]{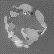}}         & Granular                 & \begin{tabular}[c]{@{}l@{}}1. Edge Ring\\ 2. Granular\end{tabular}                 \\ \hline
(b) \parbox[c]{5em}{\includegraphics[width=4em]{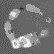}}         & Edge Ring                & \begin{tabular}[c]{@{}l@{}}1. Edge Ring\\ 2. Edge Bulge\end{tabular}               \\ \hline
(c) \parbox[c]{5em}{\includegraphics[width=4em]{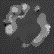}}            & Edge Bulge               & \begin{tabular}[c]{@{}l@{}}1. Edge Bulge\\ 2. Edge Ring\end{tabular}               \\ \hline
(d) \parbox[c]{5em}{\includegraphics[width=4em]{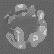}}            & Edge Bulge               & \begin{tabular}[c]{@{}l@{}}1. Edge Bulge\\ 2. Edge Ring\\ 3. Granular\end{tabular} \\ \hline
\end{tabular}
\label{tab:complextsvs}
\end{table}

\section{Conclusion}
TSV extrusion is a major reliability concern in 3D ICs, which can deform the adjacent interconnect layers and result in device failure. Recognizing the morphologies of TSV extrusion is an important step in addressing this crucial reliability concern. In this paper, we leveraged deep learning paradigms to classify TSV extrusion morphologies into three different classes based on the TSV surface profile images obtained experimentally by the WLI method. 

First, we built a framework to create a TSV extrusion morphology dataset from the raw data obtained from WLI measurement. Next, four CNN models with different network complexities were utilized for the classification task. The CNN architectures included the well-known LeNet-5 \cite{LeNet} and AlexNet \cite{AlexNet}, as well as two other CNN models proposed herein, called AlexNet-inspired LeNet and VGG-inspired AlexNet. We used data augmentation and dropout approaches to realize a balance between overfitting and underfitting in the CNN models. The results showed that the optimized VGG-inspired AlexNet model can achieve the highest classification accuracy of 86.1\% among the CNN models studied herein, which is comparable to the 89.1\% achieved by a human expert. 

Finally, the relatively high classification error of the human expert encouraged us to conduct further experiments and analyses regarding the nature of the TSV extrusion morphologies. As discussed in the last section of the paper, it was verified that in 5\%-10\% of the TSV samples more than one extrusion mechanism is involved, which led to the appearance of more than one morphology in the TSV images. This resulted in a reduction in the top-1 classification accuracy for both human experts and CNN models. This can be possibly alleviated by modifying the data labeling process and allowing the CNN models to report the appearance probability of each morphology. However, this would demand a larger dataset, which is one of the major objectives of publishing the outcome of this work. All the programs and codes developed in this project are available for the research community to contribute to the dataset, which can provide several interesting possibilities for future work.

\bibliographystyle{IEEEtran}

\balance
\bibliography{ref}

\begin{thebibliography}{10}
\providecommand{\url}[1]{#1}
\csname url@samestyle\endcsname
\providecommand{\newblock}{\relax}
\providecommand{\bibinfo}[2]{#2}
\providecommand{\BIBentrySTDinterwordspacing}{\spaceskip=0pt\relax}
\providecommand{\BIBentryALTinterwordstretchfactor}{4}
\providecommand{\BIBentryALTinterwordspacing}{\spaceskip=\fontdimen2\font plus
\BIBentryALTinterwordstretchfactor\fontdimen3\font minus
  \fontdimen4\font\relax}
\providecommand{\BIBforeignlanguage}[2]{{%
\expandafter\ifx\csname l@#1\endcsname\relax
\typeout{** WARNING: IEEEtran.bst: No hyphenation pattern has been}%
\typeout{** loaded for the language `#1'. Using the pattern for}%
\typeout{** the default language instead.}%
\else
\language=\csname l@#1\endcsname
\fi
#2}}
\providecommand{\BIBdecl}{\relax}
\BIBdecl

\bibitem{mack2011fifty}
C.~A. Mack, ``Fifty years of moore's law,'' \emph{IEEE Transactions on
  semiconductor manufacturing}, vol.~24, no.~2, pp. 202--207, 2011.

\bibitem{havemann2001integration}
R.~H. Havemann and J.~A. Hutchby, ``High-performance interconnects: An
  integration overview,'' \emph{Proceedings of the IEEE}, vol.~89, no.~5, pp.
  586--601, 2001.

\bibitem{kuhn2009moore}
K.~J. Kuhn, ``Moore's law past 32nm: Future challenges in device scaling,'' in
  \emph{2009 13th International Workshop on Computational Electronics}.\hskip
  1em plus 0.5em minus 0.4em\relax IEEE, 2009, pp. 1--6.

\bibitem{hisamotofinfet}
D.~Hisamoto, W.-C. Lee, J.~Kedzierski, H.~Takeuchi, K.~Asano, C.~Kuo,
  E.~Anderson, T.-J. King, J.~Bokor, and C.~Hu, ``Finfet-a self-aligned
  double-gate mosfet scalable to 20 nm,'' \emph{IEEE transactions on electron
  devices}, vol.~47, no.~12, pp. 2320--2325, 2000.

\bibitem{kish2002end}
L.~B. Kish, ``End of moore's law: thermal (noise) death of integration in micro
  and nano electronics,'' \emph{Physics Letters A}, vol. 305, no. 3-4, pp.
  144--149, 2002.

\bibitem{schulz1999end}
M.~Schulz, ``The end of the road for silicon?'' \emph{Nature}, vol. 399, no.
  6738, p. 729, 1999.

\bibitem{Thomasthesis}
T.~N. Theis, ``The future of interconnection technology,'' \emph{IBM Journal of
  Research and Development}, vol.~44, no.~3, pp. 379--390, 2000.

\bibitem{licata1995interconnect}
T.~J. Licata, E.~G. Colgan, J.~M. Harper, and S.~E. Luce, ``Interconnect
  fabrication processes and the development of low-cost wiring for cmos
  products,'' \emph{IBM Journal of Research and Development}, vol.~39, no.~4,
  pp. 419--435, 1995.

\bibitem{GAMBINO2015tsvfab}
J.~P. Gambino, S.~A. Adderly, and J.~U. Knickerbocker, ``An overview of
  through-silicon-via technology and manufacturing challenges,''
  \emph{Microelectronic Engineering}, vol. 135, pp. 73 -- 106, 2015.

\bibitem{HMC}
J.~Jeddeloh, ``Hybrid memory cube architecture: a closer look,'' 2016.

\bibitem{Knickerbocker2008}
J.~U. Knickerbocker, P.~S. Andry, B.~Dang, R.~R. Horton, M.~J. Interrante,
  C.~S. Patel, R.~J. Polastre, K.~Sakuma, R.~Sirdeshmukh, E.~J. Sprogis, S.~M.
  Sri-Jayantha, A.~M. Stephens, A.~W. Topol, C.~K. Tsang, B.~C. Webb, and S.~L.
  Wright, ``Three-dimensional silicon integration,'' \emph{IBM Journal of
  Research and Development}, vol.~52, no.~6, pp. 553--569, Nov 2008.

\bibitem{Wolf2008hetero}
M.~J. {Wolf}, P.~{Ramm}, A.~{Klumpp}, and H.~{Reichl}, ``Technologies for
  \uppercase{3D} wafer level heterogeneous integration,'' in \emph{2008
  Symposium on Design, Test, Integration and Packaging of MEMS/MOEMS}, 2008,
  pp. 123--126.

\bibitem{lau2010evolution}
J.~H. Lau, ``Evolution and outlook of \uppercase{TSV} and \uppercase{3D IC/S}i
  integration,'' in \emph{2010 12th Electronics Packaging Technology
  Conference}.\hskip 1em plus 0.5em minus 0.4em\relax IEEE, 2010, pp. 560--570.

\bibitem{garrou2014handbook}
P.~Garrou, M.~Koyanagi, and P.~Ramm, \emph{Handbook of 3D Integration}.\hskip
  1em plus 0.5em minus 0.4em\relax John Wiley \& Sons, 2014.

\bibitem{Ryu2011}
S.~Ryu, K.~Lu, X.~Zhang, J.~Im, P.~S. Ho, and R.~Huang, ``Impact of
  near-surface thermal stresses on interfacial reliability of through-silicon
  vias for \uppercase{3-D} interconnects,'' \emph{IEEE Transactions on Device
  and Materials Reliability}, vol.~11, no.~1, pp. 35--43, March 2011.

\bibitem{Rangan2008}
N.~Ranganathan, K.~Prasad, N.~Balasubramanian, and K.~L. Pey, ``A study of
  thermo-mechanical stress and its impact on through-silicon vias,''
  \emph{Journal of Micromechanics and Microengineering}, vol.~18, no.~7, p.
  075018, 2008.

\bibitem{slevan2009}
C.~S. Selvanayagam, J.~H. Lau, X.~Zhang, S.~K.~W. Seah, K.~Vaidyanathan, and
  T.~C. Chai, ``Nonlinear thermal stress/strain analyses of copper filled
  \uppercase{TSV} (through silicon via) and their flip-chip microbumps,''
  \emph{IEEE Transactions on Advanced Packaging}, vol.~32, no.~4, pp. 720--728,
  Nov 2009.

\bibitem{jiang2015MRS}
T.~Jiang, J.~Im, R.~Huang, and P.~S. Ho, ``Through-silicon via stress
  characteristics and reliability impact on \uppercase{3D} integrated
  circuits,'' \emph{MRS Bulletin}, vol.~40, no.~3, pp. 248–--256, 2015.

\bibitem{Jalilvand2017}
G.~{Jalilvand}, O.~{Ahmed}, K.~{Bosworth}, C.~{Fitzgerald}, Z.~{Pei}, and
  T.~{Jaing}, ``Application of a metallic cap layer to control \uppercase{C}u
  \uppercase{TSV} extrusion,'' in \emph{2017 IEEE 67th Electronic Components
  and Technology Conference (ECTC)}, 2017, pp. 61--66.

\bibitem{jalilvand2019ectc}
G.~Jalilvand and T.~Jiang, ``Study of the effect and mechanism of a cap layer
  in controlling the statistical variation of via extrusion,'' in \emph{2019
  IEEE 69th Electronic Components and Technology Conference (ECTC)}.\hskip 1em
  plus 0.5em minus 0.4em\relax IEEE, 2019, pp. 1909--1915.

\bibitem{JALILVAND2019scripta}
G.~Jalilvand, O.~Ahmed, L.~Spinella, L.~Zhou, and T.~Jiang, ``The effective
  control of \uppercase{C}u through-silicon via extrusion for three-dimensional
  integrated circuits by a metallic cap layer,'' \emph{Scripta Materialia},
  vol. 164, pp. 101 -- 104, 2019.

\bibitem{DeMessemaeker2014}
J.~D. Messemaeker, O.~V. Pedreira, H.~Philipsen, E.~Beyne, I.~D. Wolf, T.~V.
  der Donck, and K.~Croes, ``Correlation between \uppercase{C}u microstructure
  and \uppercase{TSV C}u pumping,'' in \emph{2014 IEEE 64th Electronic
  Components and Technology Conference (ECTC)}, May 2014, pp. 613--619.

\bibitem{jalilvand2020ectc}
G.~{Jalilvand}, O.~{Ahmed}, N.~{Dube}, and T.~{Jiang}, ``Study of the impact of
  pitch distance on the statistical variation of tsv protrusion and the
  underlying mechanisms,'' in \emph{2020 IEEE 70th Electronic Components and
  Technology Conference (ECTC)}, 2020 (\textit{in print}).

\bibitem{JALILVAND2020IEEECPMT}
G.~Jalilvand, O.~Ahmed, N.~Dube, and T.~Jiang, ``The effect of pitch distance
  on the statistics and morphology of through-silicon via extrusion,''
  \emph{IEEE Transactions on Components, Packaging and Manufacturing
  Technology}, 2020 (under review).

\bibitem{Gwyddion}
D.~Nečas and P.~Klapetek, ``Gwyddion: an open-source software for spm data
  analysis,'' \emph{Open Physics}, vol.~10, no.~1, pp. 181 -- 188, 2012.

\bibitem{dataaugmentationsurvey}
C.~Shorten and T.~M. Khoshgoftaar, ``A survey on image data augmentation for
  deep learning,'' \emph{Journal of Big Data}, vol.~6, no.~1, p.~60, 2019.

\bibitem{LeNet}
Y.~{Lecun}, L.~{Bottou}, Y.~{Bengio}, and P.~{Haffner}, ``Gradient-based
  learning applied to document recognition,'' \emph{Proceedings of the IEEE},
  vol.~86, no.~11, pp. 2278--2324, 1998.

\bibitem{AlexNet}
A.~Krizhevsky, I.~Sutskever, and G.~E. Hinton, ``Imagenet classification with
  deep convolutional neural networks,'' in \emph{Advances in Neural Information
  Processing Systems 25}, F.~Pereira, C.~J.~C. Burges, L.~Bottou, and K.~Q.
  Weinberger, Eds.\hskip 1em plus 0.5em minus 0.4em\relax Curran Associates,
  Inc., 2012, pp. 1097--1105.

\bibitem{VGG-16}
K.~Simonyan and A.~Zisserman, ``Very deep convolutional networks for
  large-scale image recognition,'' \emph{arXiv preprint arXiv:1409.1556}, 2014.

\bibitem{tensorflow}
M.~Abadi, P.~Barham, J.~Chen, Z.~Chen, A.~Davis \emph{et~al.}, ``Tensorflow: A
  system for large-scale machine learning,'' in \emph{12th {USENIX} Symposium
  on Operating Systems Design and Implementation ({OSDI} 16)}.\hskip 1em plus
  0.5em minus 0.4em\relax Savannah, GA: {USENIX} Association, Nov. 2016, pp.
  265--283.

\bibitem{keras}
\BIBentryALTinterwordspacing
F.~Chollet \emph{et~al.} (2015) Keras. [Online]. Available:
  \url{https://github.com/fchollet/keras}
\BIBentrySTDinterwordspacing

\bibitem{dropout}
N.~Srivastava, G.~Hinton, A.~Krizhevsky, I.~Sutskever, and R.~Salakhutdinov,
  ``Dropout: A simple way to prevent neural networks from overfitting,''
  \emph{Journal of Machine Learning Research}, vol.~15, no.~1, p. 1929–1958,
  2014.

\end{thebibliography}
%



\end{document}